\documentclass{sig-alternate}

\usepackage{hyperref}
\usepackage{textcomp}

\begin{document}

\title{Video Manipulation Techniques for the Protection of Privacy in Remote Presence Systems}

\numberofauthors{2} 
\author{
\alignauthor Alexander Hubers, Emily Andrulis, Levi Scott, Tanner Stirrat, Duc Tran, Ruonan Zhang, Ross Sowell\\
       \affaddr{Department of Computer Science}\\
       \affaddr{Cornell College}\\
       \affaddr{Mount Vernon, IA, USA}
       \footnotemark[1]
\alignauthor Cindy Grimm, William D. Smart\\
       \affaddr{School of MIME}\\
       \affaddr{Oregon State University}\\
       \affaddr{Corvallis, OR, USA}
       \footnotemark[2]
}
\maketitle

\begin{abstract}
Systems that give control of a mobile robot to a remote user raise privacy concerns about what the remote user can see and do through the robot.  We aim to preserve some of that privacy by manipulating the video data that the remote user sees.  Through two user studies, we explore the effectiveness of different video manipulation techniques at providing different types of privacy.  We simultaneously examine task performance in the presence of privacy protection.  In the first study, participants were asked to watch a video captured by a robot exploring an office environment and to complete a series of observational tasks under differing video manipulation conditions.   Our results show that using manipulations of the video stream can lead to fewer privacy violations for different privacy types. 
Through a second user study, it was demonstrated that these privacy-protecting techniques were effective without diminishing the task performance of the remote user.
\end{abstract}




\begin{figure}[h!]
\centering
\includegraphics[width=1.0\linewidth, scale=.5]{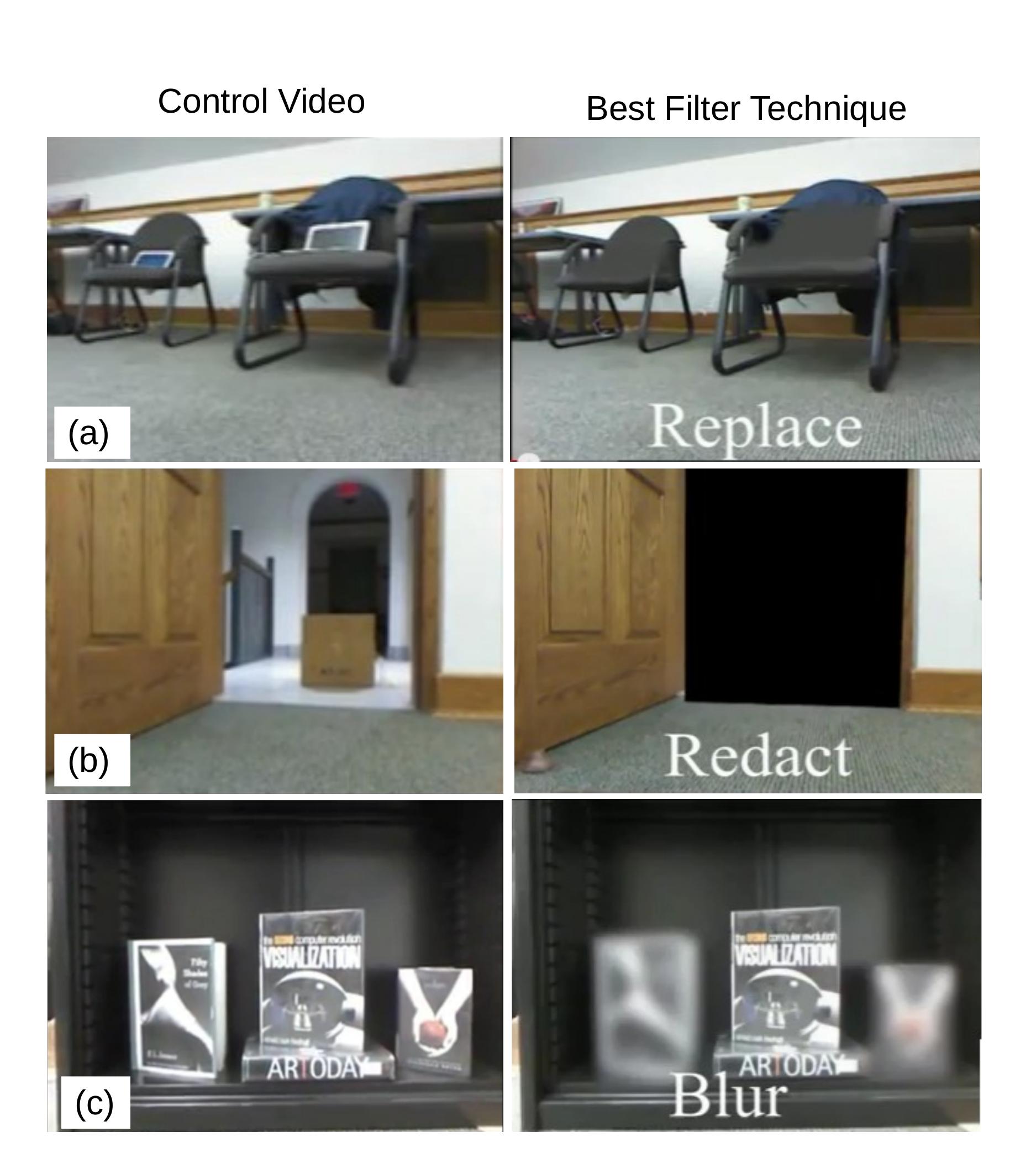}
\caption[Best Filters]{Unfiltered video compared to best filter techniques based on desired privacy type. A replace filter is applied to two laptops so that the user ``Can't tell'' they are there (a). A redact filter is used so that the user ``Can't observe'' anything in the hallway (b). A blur filter is placed over two books so that a user ``Can't discern'' their titles (c).}
\label{fig:best}
\end{figure}

\section{Introduction}
What makes a mobile remote presence system more of a privacy concern than, say a Skype connection?  The main difference is a shift in control.  If you do not want someone to see you in the shower over Skype, then you do not take your laptop into the bathroom.  You have control over what the user on the other end sees.  However, with mobile remote presence systems, the remote user can pilot the system around the world, giving them control over what they see.

\footnotetext[1]{\{ahubers15, eandrulis16, lscott15, tstirrat15, rzhang16, and rsowell\}@cornellcollege.edu}
\footnotetext[2]{\{cindy.grimm,bill.smart\}@oregonstate.edu}
For the purposes of this paper, we define a remote presence system (RPS) to be a system that allows a remote operator to be virtually present in another location, and to interact with the people and things there.  The telephone and Skype are two common examples of such systems.  However, the focus of the work in this paper is on systems that go beyond just observing their environment, but also allow the user to act in, and on it.  The canonical example of such a system is the Personal Roving Presence (PRoP) project~\cite{Paulos:2001}, a mobile robot base mounting an LCD screen, camera, and speaker on a human-height pole.  The robot wandered around the UC Berkeley campus and surrounding areas under the control of a remote operator, allowing the operator to interact with people the robot encountered.  A decade later, similar systems are becoming commercially available, including the Beam~\cite{SuitableTech:2012}, VGo ~\cite{VGo:2012}, and InTouch RP-VITA~\cite{InTouch:2012}.  The caricature of these systems is that they are ``Skype on a stick'':  a traditional video-conference interaction where the user has some control over the physical location of the system.

These mobile systems introduce new privacy concerns.  With a passive teleconference system, the remote participant can only look at what the proximal participant points the camera at.  On mobile systems, the remote operator can now point the camera themselves, taking control of privacy away from the proximal user.

One method of preserving the privacy of the proximal user is to alter the visual data that the  remote operator sees. The central question that we investigate in this paper is:  What methods of video stream manipulation are most effective at ensuring a given privacy type without diminishing operator performance? Our specific contributions are listed below.

\paragraph{Contributions}
\begin{enumerate}
  \item Define different types of privacy based on what can be observed (Can't tell, Can't observe, Can't discern).
  \item Perform a user study to determine which video manipulation techniques (redact, blur, replace, abstract) works best for a given privacy type.
	\item Demonstrate that the remote user can still perform tasks with the privacy video manipulations in place.
\end{enumerate}

\paragraph{Overview}
We first define the privacy types and video manipulation techniques and summarize related work. Section~\ref{sec:vidManipStudy} describes our first user study examining the effectiveness of video manipulation techniques at ensuring different privacy types. A second study that investigates the impact of these privacy-protecting measures on task performance is presented in Section~\ref{sec:taskPerfStudy}. Section~\ref{sec:conclusion} concludes with a brief review of the two studies and their key results.     

\begin{figure*}
\centering
\includegraphics[width=1.0\linewidth]{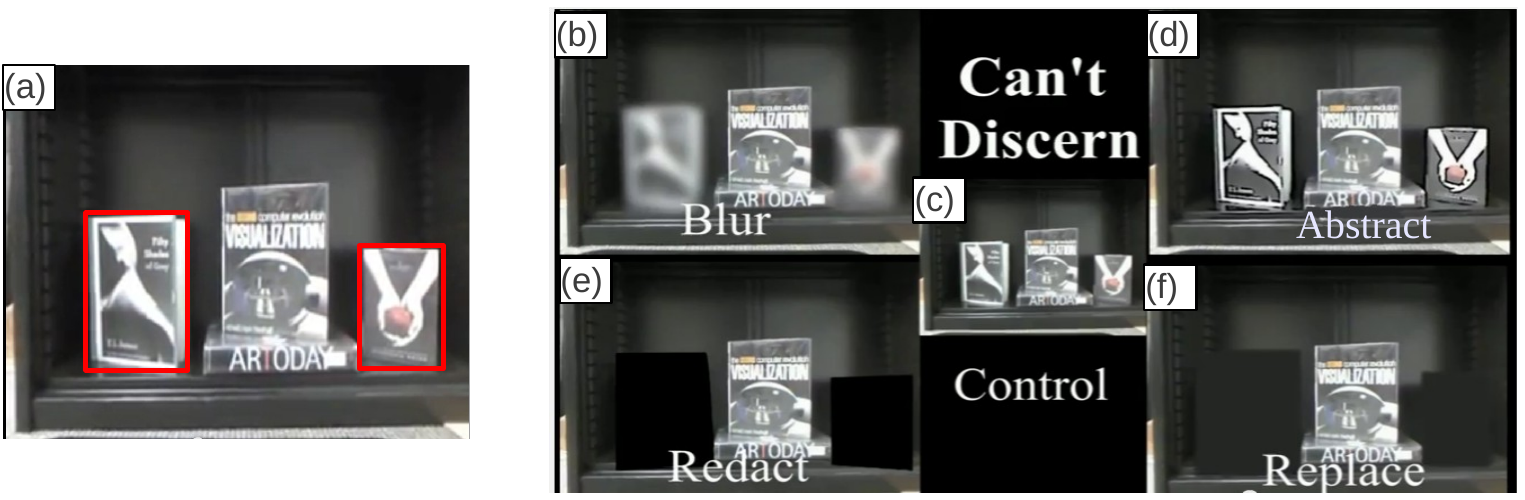}
\caption[Video manipulation techniques]{Video manipulation techniques.  Left:  The original image is shown with the two book titles to be filtered highlighted in red (a).  Right:  Four different video manipulation techniques are applied to the book titles --- Blur (b),  Control (c), Abstract (d), Redact (e), and Replace (f).}
\label{fig:filters}
\end{figure*}

\subsection{Observational Privacy Types}
We define a \textit{privacy type} to be a specific restriction on the capabilities of the remote presence system.  The capabilities can either be \textit{physical}, where the system is prevented from taking some action (the RPS cannot enter/leave some area or touch some object), or \textit{observational}, where the sensor data transmitted to the remote user are censored in some way.
 Sensor data can be both visual or auditory, but we limit our scope to visual data due to the robot we are using.  In this paper, we focus on techniques for ensuring different types of observational privacy as related to the video stream, since this is both the most challenging sensor modality and also the one most likely to violate privacy.  

\paragraph{Defined Privacy Types}
We consider implementations for three types of observational privacy:  ``Can't tell'', ``Can't discern'', and ``Can't observe.''  We frame these as restrictions on the remote user, rather than on the RPS itself.  The RPS has unrestricted access to the data from its sensors, but only passes on a modified version of these data to the remote user. 

\begin{enumerate}
  \item \textbf{Can't tell}.  The expectation is that the remote operator cannot tell if a particular object is there or not. Examples include:  Not noticing any items exist on a table, unable to tell that there is a person present in the room.  
  \item \textbf{Can't observe}.  The expectation is that the remote operator might be able to tell there is something there, but cannot directly perceive it.  Examples include:  Not being able to look into a certain room, not being able to identify a shape as a person, or not showing what types of objects are on a table.
  \item \textbf{Can't discern}.   The expectation is that the remote operator can tell that there is something there and can identify the class of the object, but not the particular instance.  Examples include:  Unable to read the text of documents on the table, unable to make out facial features, and unable to make out details of pictures on the walls.
  
\end{enumerate}

\subsection{Video Manipulation Techniques}
In order to preserve privacy we need to alter the visual data the operator is seeing.  There are a variety of methods for doing so, many of which arise from the field of non-photorealistic or artistic rendering.  We discuss related work in that area here, and then we list the specific techniques that we evaluate for privacy protection in this paper.

Broadly speaking, we classify manipulation of images and video by how they change the image:  blur, inpainting, abstraction, line drawings, and painterly rendering.  Blurring is a straightforward image filter and is commonly used in TV to obscure people's faces.  Inpainting~\cite{Barnes:2009, Bugeau:2010, Drori:2003, Herling:2012, Sun:2005, VijayVenkatesh:2009} allows for filling an area of an image with synthesized content that is ideally indistinguishable from its environment.  Abstraction, also sometimes called image stylization~\cite{Kyprianidis:2011, Winnemoller:2012, Mould:2012, Winnemoller:2011, DeCarlo:2002}, is similar to blurring, in that details are elided, but it differs in that strong edges are preserved.  It can also involve restricting the color palate to create a cartoon-like effect.  Since these are essentially texture filters, most methods can be efficiently implemented on a GPU~\cite{Winnemoller:2006, Zhao:2009}.  Line drawings~\cite{DeCarlo:2003, Rusinkiewicz:2005, Eisemann:2008, Gooch:2002, Gooch:2004} similarly preserve edges, but eliminate color information and render the result as a pen and ink or pencil-style sketch (sometimes with shading represented as hatching~\cite{Singh:2010}).  Painterly rendering techniques try to mimic a particular style, such as pixelation~\cite{Gerstner:2012}, oil or watercolors~\cite{Lu:2010, Olsen:2011, Xu:2005, Rosin:2010} and comic-style~\cite{Sauvaget:2008}.  Although not always intentional, most of these techniques also result in some image simplification or loss of detail, especially with large brush sizes.

One primary concern when working with video or a moving camera is that the image can flicker because the lines and strokes change over time.  There are several methods for addressing this~\cite{Rusinkiewicz:2005, Lu:2010}, with the primary approach being to evolve the current stylization to the next frame, rather than starting from scratch.  

\paragraph{Defined Video Manipulations}
For our initial study we have chosen four basic video manipulation techniques to evaluate with respect to their effectiveness at providing different types of observational privacy (see Figure~\ref{fig:filters}).  We will refer to them as redact, blur, replace, and abstract.  These were chosen because they are representative of the different classes of techniques that we might apply to protect privacy, implementations were readily available, and they can be efficiently implemented on a GPU to achieve interactive rates. 

\begin{enumerate}
  \item \textbf{Redact}.  We hide something by removing it from the video stream, i.e., blacking it out.
  \item \textbf{Blur}.  We obscure a specified portion of the video data by applying a Gaussian blur.
  \item \textbf{Replace}.  We replace an object with color data from its surrounding environment (in-painting).
  \item \textbf{Abstract}.  We abstract a portion of the video by applying a combination of bilateral and meanshift filtering.
\end{enumerate}

\subsection{Video Manipulation and Privacy}
Various studies have looked into how using video manipulations may help uphold privacy. Specifically, privacy typically considers autonomy, confidentiality, and solitude~\cite{Boyle:2009}. Filtering out parts of an image through marker detection has been shown to effectively uphold privacy for video surveillance cameras~\cite{Schiff:2007}. With an always-on camera space, using a blur filter has been shown to better balance protecting one's privacy while still allowing sufficient awareness to the user, so that any necessary and relevant information may still be gleaned from the image both with a co-present media space~\cite{Kim:2007} and a telepresent media space~\cite{Boyle:2000}. However, in some circumstances where the privacy concerns are greater (i.e. assistive monitoring through use of a fixed always-on camera), a blur filter may not be sufficient, and another technique such as redact may work more effectively~\cite{Edgcomb:2012}. 

In our study we examine which video manipulation techniques work best for the three different privacy types we have defined in this section, while still allowing the user to complete a given task. Although research has been done on the effects of using video manipulations to uphold privacy, these studies have been conducted for use in video media spaces that have a fixed camera. In our study, we investigate to what degree these video manipulations uphold privacy when the camera is capable of traversing around a room and examining objects from different angles of view. 

\subsection{Video Manipulation and Task Performance}
Several studies have looked at the effect of abstraction or image simplification on task performance, perception, and affect.  In general, abstraction tends to speed up object recognition and visual search tasks~\cite{Gooch:2004, Redmond:2009, Fischer:2006}.  However, there is evidence that increased abstraction can lead to less accurate predictions of shape~\cite{Fischer:2006, Wallraven:2007, Redmond:2009, Schumann:1996} and a reduced emotional or believability response~\cite{Duke:2003, Halper:2003, Schumann:1996, Mould:2012:SSC}.  There is also evidence that people routinely underestimate distances in virtual environments, and that abstraction can exacerbate this~\cite{Gooch:2002, Phillips:2009}.  Selective abstraction is also effective at focusing a viewer on the image point of interest by eliding detail in unimportant areas of the image~\cite{Cole:2006}.

\subsection{Hypotheses}
In the studies presented in this paper, our goal is to determine which methods of video manipulation are most effective at ensuring different types of observational privacy without interfering with task performance.  To evaluate the video manipulation techniques, we test the following specific hypotheses:

\begin{description}
\item[H1.]  Based on the desired privacy type, selected video manipulation techniques lead to fewer violations of privacy expectations.
\item[H2.]  Using the privacy video manipulation technique does not lead to a lowered task performance.
\end{description}

\section{Video Manipulaton Study}
\label{sec:vidManipStudy}
In this study we tested how well the different video manipulation techniques worked for each of the different privacy types, using three different scenarios.  To avoid issues with localization, tracking, and training users to drive the robot, we conducted this study with videos captured from the robot's camera.

\subsection{Method}
Participants were asked to watch three short video clips that were captured by a robot exploring an office environment, and to respond to five questions asking them to identify objects within the environment. Each scene had a specific privacy type applied to it, and participants viewed one clip of each scene. Each clip had one of the five randomly assigned video manipulations applied to it.


\subsubsection{Participants}
140 participants were recruited through Amazon's Mechanical Turk. Participants were compensated between 20 and 40\textcent  
\: for their participation. Participants were told that they would be expected to ``watch a clip from the perspective of a robot investigating an office and answer 5 short questions.'' The average time spent per participant was between 3-5 minutes.

\subsubsection{Procedure}
Participants were given the following prompt:

\begin{quote}
A mobile robot has explored an office environment for you and acquired the videos on the following three pages. When you are ready to begin, please click ``Continue to Videos" below to watch the video and then answer the following questions about what you saw. The questions will be divided into three pages, each containing a separate video. You may take notes, and you may pause, rewind, or replay the video as often as you like. However, once you begin the test, you may not exit out and come back, return to the previous page, or refresh to repeat it.
\end{quote}

Each page provided a video clip of a different scene that used one randomly assigned video manipulation technique. By the end, each participant had viewed one video clip from each of the three scenes and each clip's video manipulation had been randomly assigned.

 The video clips were followed by questions that asked the participant to identify objects in the video. Each scene had a different set of questions, based on the content of the scene. The answers participants gave in response to those questions in no way affected other scenes or questions in other scenes.


\subsubsection{Control Videos and Privacy Expectations}
\label{sec:controlAndExpect}
The videos used in our experiment were captured from the video stream of the Kinect sensor on a TurtleBot 2~\cite{TurtleBot:2012} that was navigated via teleoperation through a staged office environment. Care was taken to ensure that all objects relevant to the tasks were clearly visible at some point during the videos.  All videos used in our study are included in the supplementary materials. 

\begin{figure}[h]
\centering
\includegraphics[width=1.0\linewidth]{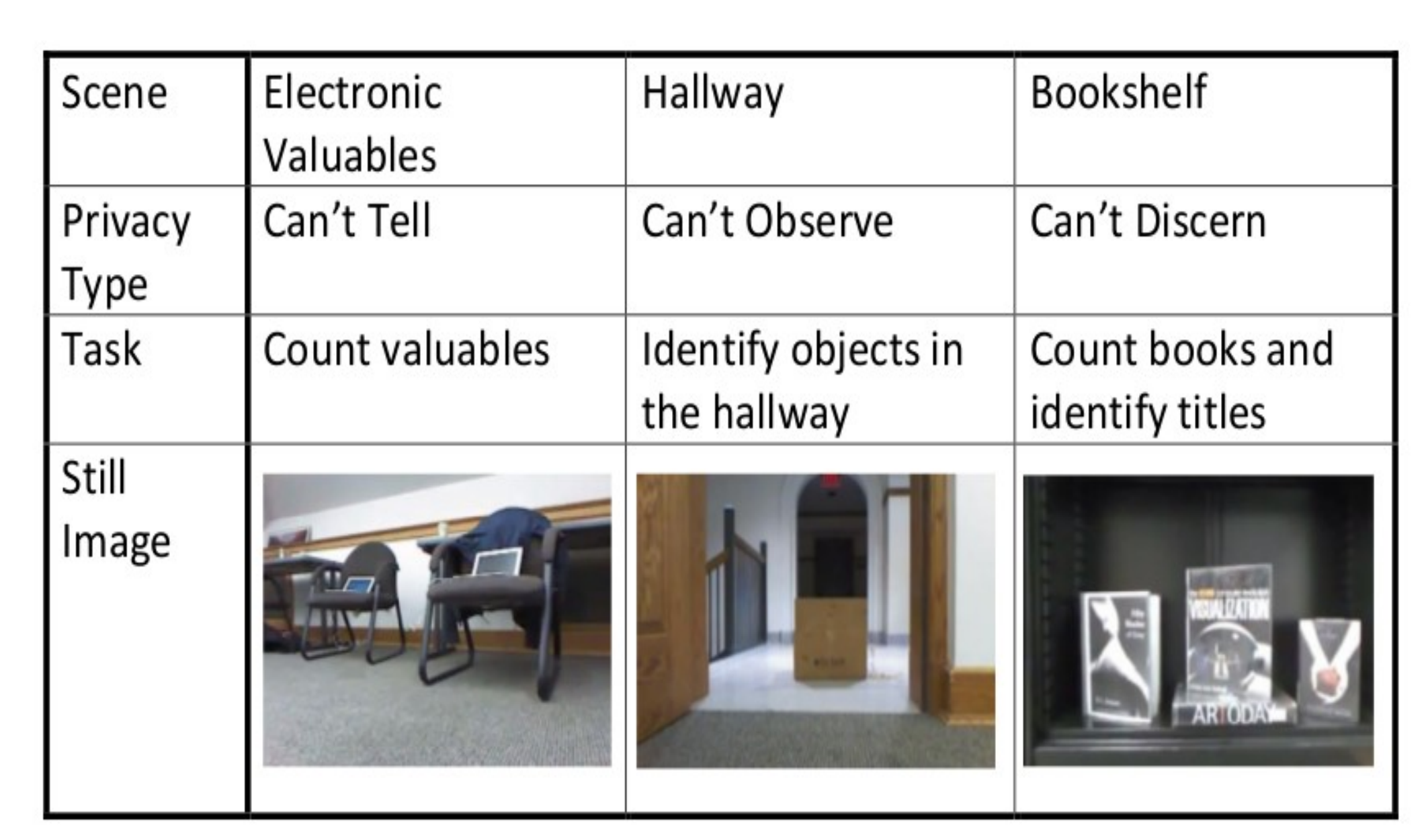}
\caption[Scenes]{This table shows the three different scenes and their corresponding privacy types, tasks, and a still image from the control video.}
\label{fig:scenes}
\end{figure}

Three different video clips were designed to provide an example of why each type of privacy might be pertinent. Figure~\ref{fig:scenes}  gives an overview of the privacy type and task for each scene as well as a still image from the control video. We describe each video scene below.
\begin{enumerate}
  \item\textbf{Electronic Valuables}. This video clip explored the office environment, passing some valuable items along the way.  A ``Can't tell'' privacy type was applied to the electronic valuables in this scene. In order of appearance, the valuables in the scene were a flat screen television on a shelf, a desktop computer monitor in the corner, and two ASUS laptops on chairs in the middle of the room.  For this scene, the proximal user has specified that the remote user should not be able to correctly identify the number of valuable items, but instead should report viewing zero total valuables. Additionally, the remote user should not be able to identify the locations of any valuable items nor any details about them. In short, the remote user should not be able to tell that these items exist in the room. 

\underline{Survey Questions}
  \begin{itemize} 
    \item ``How many computers or electronic valuables (costing more than \$50) are there in the room?'' 
    \item  ``Name the valuables and the their locations in the room.''
  \end{itemize}
  \item\textbf{Hallway}. This video clip recorded viewing out a door into a hallway that contained one cardboard box in the center.  A ``Can't observe'' privacy type was applied to this scene. In this clip the proximal user has specified that the user must not be able to see outside the room, so the user was not allowed to see anything past the open doorway into the hall. Although the cardboard box was the object of interest, it was expected that the user would not be able to see the cardboard box nor anything else in the hallway. 

\underline{Survey Questions}
  \begin{itemize} 
    \item  ``Could you see anything in the hallway? If so, please list anything that you saw.''
  \end{itemize}
  \item\textbf{Bookshelf}. This video clip recorded viewing an office bookshelf, and particularly, the four books on the bookshelf.  A ``Can't discern'' privacy type was applied to this scene. These four books on the bookshelf were ``ARToday,'' ``Visualization: The Second Computer Revolution,'' ``Fifty Shades of Grey,'' and ``Twilight.'' For this privacy expectation the proximal user does not mind that the books are visible, but does not want to reveal an affinity for embarrassing romance novels. For this reason, the remote user is expected to identify correctly that there are four books, but not be able to read the titles of ``Twilight'' or ``Fifty Shades of Grey.''  However, the user should still be able to identify ``ARToday'' and ``Visualization: The Second Computer Revolution.''

 \underline{Survey Questions}
  \begin{itemize} 
    \item ``How many books are on the bookshelf?''
    \item ``Name as many titles of the books as you can.''
  \end{itemize}
\end{enumerate}

\subsubsection{Video Manipulation Independent Variable}
All participants watched a video clip from each of the three scenes, one for every privacy type. Each of the clips was randomly assigned from the five video manipulations that were applied to every scene. A total of 15 videos were created: five distinct video manipulations (Redact, Replace, Abstract, Blur, and the control) applied to each of the three scenes (Electronic Valuables, Hallway, and Bookshelf). In the manipulated video clips the objects that were filtered were the four valuables (two laptops, one flat screen television, and one desktop computer) in the Electronic Valuables scene, the doorway in the Hallway scene, and the two books (``Twilight" and ``Fifty Shades of Grey") in the Bookshelf scene. Although multiple items may have been filtered in one scene, all the filters were of the same type of video manipulation within a single video clip. These video manipulations were all applied using Adobe After Effects\textsuperscript{\textregistered} on the original control video.

\subsubsection{Measures}
Our primary measurement is the participants' responses to the survey questions.  We also record the time that it took for participants to answer all questions before going on to the next page. We then coded the recorded answers based on the privacy type.

\paragraph{Privacy Violations}
We measure the degree to which privacy was protected in a distinct manner for each privacy condition, based on the expectations for that type of privacy as described in Section~\ref{sec:controlAndExpect}.  For the Electronic Valuables scene, one privacy violation was coded for each correctly identified valuable. In the Hallway scene, the privacy violation was recorded if the participant saw anything at all in the hallway. With the Bookshelf scene, one privacy violation was coded for each correctly identified filtered title (i.e. ``Twilight'' or ``Fifty Shades of Grey''). Although we had two separate researchers code the data, the results presented are based on the principle coder's findings. More detailed information on our coding scheme and inter-rater reliability is available in the supplemental materials.

\paragraph{Additional Expectations}
Additionally, the user was expected to be able to identify the number of books and the titles of the two unflitered books on the book shelf.  With this privacy type the participant should have recorded four books as the number of books on the shelf. Participants were also expected to correctly identify two book titles, ``ARToday'' and ``Visualization: The Second Computer Revolution.''

\begin{figure*}
\centering
\includegraphics[width=1.0\linewidth]{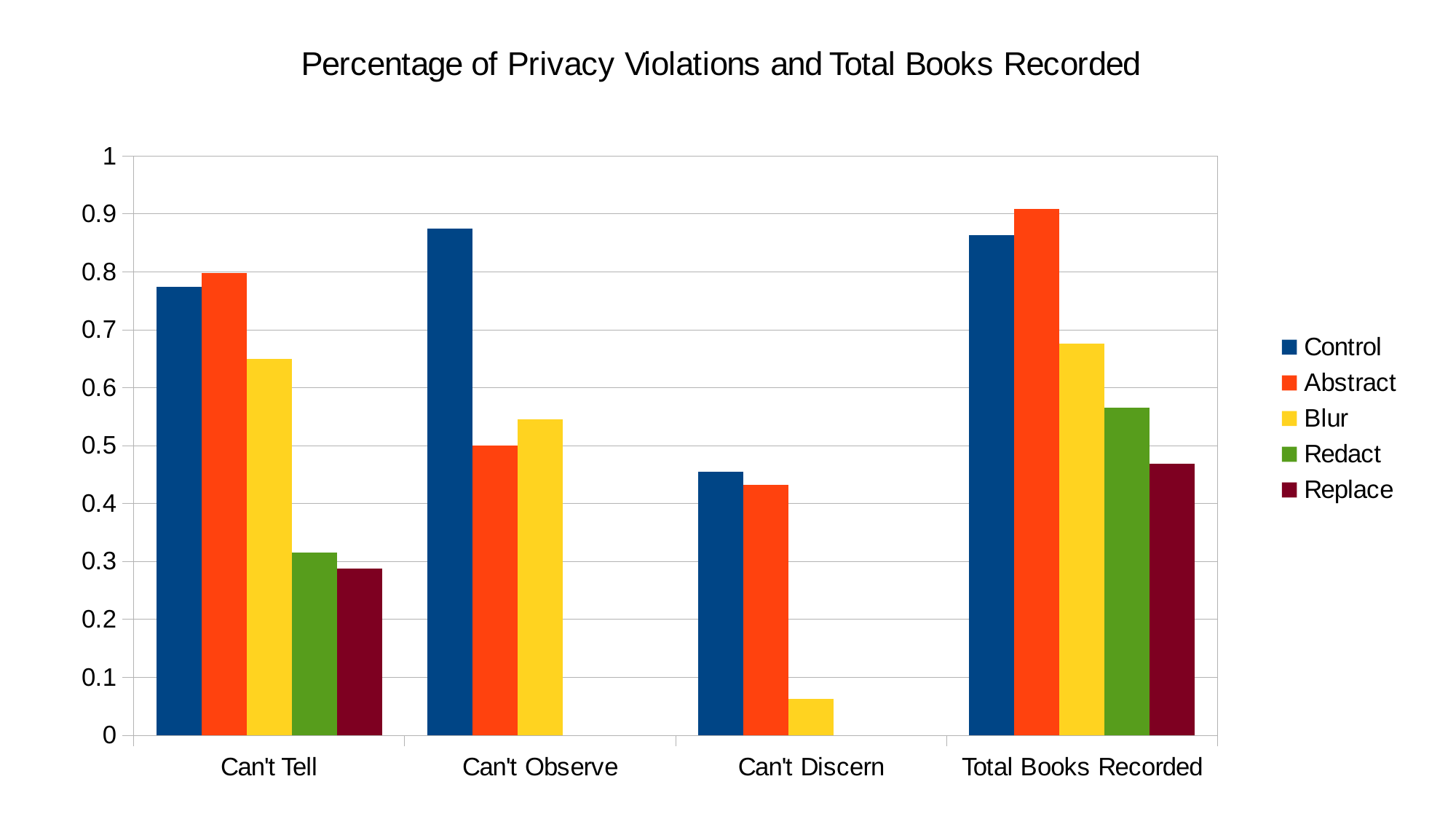}
\caption[Violations Results]{Results for privacy violations and total books recorded. The first three charts show the average percentage of privacy violations that occurred for each video manipulation. Each privacy type had a different number of possible privacy violations: four for ``Can't tell,'' one for ``Can't observe,'' and two for ``Can't discern.'' The data shown illustrates the percentage of total possible privacy violations that occurred on average for each video manipulation. The ideal percentage of privacy violations is always zero. The last graph shows the average percentage of total books that were recorded for each video manipulation during the Bookshelf scene. There were four books total, and the ideal percentage of total books recorded is 100\%.}
\label{fig:violations}
\end{figure*}

\begin{figure*}
\centering
\includegraphics[width=1.0\linewidth]{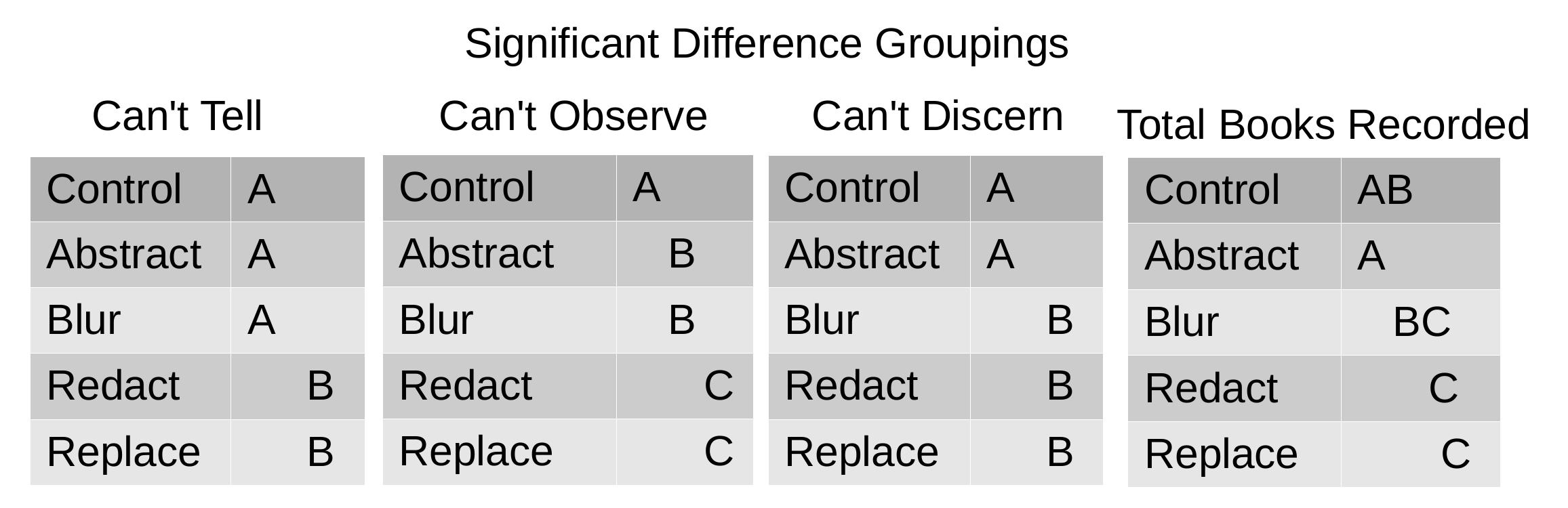}
\caption[Significant Difference Groupings]{Significant Difference Groupings. Using Tukey's method for multiple comparisons with a 95\% confidence interval, these tables show the groupings showing significant differences. Video manipulations within the same letter group are not significantly different from each other, while video manipulations with no shared letter groupings are significantly different from each other. Video manipulations with two letter groupings are not significantly different from either of the two letter groups they belong to.}
\label{fig:sigdif}
\end{figure*}

\subsection{Results}
\subsubsection{Privacy Violations}
The privacy violations were analyzed and recorded in different ways for each privacy type, as explained above. Figure~\ref{fig:violations} shows the percentage of privacy violations that were recorded on average for each privacy type based on the video manipulation, and it shows the percentage of the total books that participants recorded seeing based on video manipulation. ANOVAs were used for each privacy type to determine if there were any significant differences between the different video manipulations, and from there Tukey's method was used for multiple comparisons with a 95\% confidence interval to determine which differences between video manipulations were significant (see Figure~\ref{fig:sigdif}). 

\paragraph{``Can't tell''}
Abstract, control and blur had the highest average percentage of privacy violations for the ``Can't tell'' privacy condition, and the difference between these three was not found to be significant. However, the drop in privacy violations between the blur and redact video manipulations was significant. Redact and replace had the lowest percentage of privacy violations on average, and the difference between redact and replace was not significant.

\paragraph{``Can't observe''}
The control video had the highest percentage of privacy violations, and the difference between this and blur was significant. No significant different was found between the blur and abstract video manipulations. There was a significant difference though between the blur and redact video manipulations. Both redact and remove video manipulations had the least amount of privacy violations with zero recorded privacy violations, so no difference was found between them.

\paragraph{``Can't discern''}
Although there were two possible privacy violations for the ``Can't discern'' privacy type, none of the video manipulations averaged more than one privacy violation. The control and abstraction video manipulation had the highest percentage of total possible privacy violations and no significant difference was found between the two. The blur video manipulation had only 6.25\% of the total possible privacy violations accruing on average, and although redact and replace recorded no privacy violations at all, there was no significant difference found between these three video manipulations. All three proved to work well to diminish privacy violations.

\subsubsection{Additional Expectations}
The ``Can't discern'' privacy type also had expectations aside from simply recording privacy violations. The results for counting the number of books are shown in Figure~\ref{fig:violations}.  The correct number of books on the bookshelf was four. Users who saw the abstracted video recorded seeing 91\% of the total books on average, which was the closest any group came to seeing all the books on the bookshelf. There was no significant difference between this and the control video. However, there was a significant difference between the control and the blur video manipulations. The differences between the blur and redact video manipulations and the redact and replace video manipulations were not significant, yet there was a significant difference between the blur and replace video manipulations.   

Accuracy in identifying the two unflitered books titles was analyzed, but no significant difference was found between the different video manipulations on correctly identifying the two titles ($F_{4,110} = 1.54, p > .05$). 
Therefore this measurement cannot be used to conclude in any way that one video manipulation is better than another in upholding this expectation of the ``Can't discern'' privacy type.

\subsubsection{Time}
Although time spent was analyzed, no significant information was revealed in analyzing the data through ANOVA. The difference in time spent for each privacy condition among the different video manipulations was insignificant ($F_{4,100} = 1.94, p > .05$) 
and the standard deviations for these average times were disproportionately large. Therefore, it was concluded that no significance could be gathered from analyzing the recorded times it took participants to complete each section of the study. 

\subsection{Analysis}
The data we collected supports our hypothesis that the given privacy types can be upheld through the use of video manipulations (H1).  In analyzing the results we discuss each privacy type separately to better consider the full significance of all the recorded results. 

\subsubsection{Can't Tell}
 A fair number of privacy violations were recorded even for the redact and replace filters, which were the two video manipulation techniques with the lowest averages. There were many factors that could have led participants to correctly identify more valuables in the scene, but the two most significant were that the environment made it easy to infer the identification of valuables and the valuables were not well defined. Although the valuable items themselves may have been filtered, other related items (e.g. cords from laptops, CPU attached to desktop computer monitor, speakers surrounding television) were not filtered and could be clearly identified around the filtered areas. The other issue was that there seemed to be confusion on what the valuables to be recorded really were, and participants would often record objects from the scene that had not been meant to be labeled as electronic valuables. This was true across the different video manipulations, including those who had viewed the control clip. The fact that control also encountered this problem shows that it was a flaw in the study design and not an effect of the filters. For these reason, the number of total valuables recorded was not analyzed.

Although one might hypothesize that replace would be significantly more useful in hiding an object completely from the screen and making it seem as though the object was never there, the data does not support that, since no significant difference was found between replace and redact video manipulations. Despite this, we concluded that the replace video manipulation had best upheld this privacy type and its expectations because in their responses, participants never made mention that something was being taken out of the picture. However, with the redact video manipulation, it was obvious from some responses that the black boxes alerted the participant that something was there, even if it was unclear what that object was. A snippet from a typical response exemplifies this problem: ``it looked like 2 laptops on chairs blocked by black  squares.'' We did not have any questions to tease out how many participants noticed that there were objects being filtered out, so all we can rely on are the free-form responses. From those answers, we can conclude that replace would be a better choice for the ``Can't tell'' privacy type than redact.

\subsubsection{Can't Observe}
Redact and replace completely upheld this privacy expectation since neither allowed any user to describe anything in the hallway. While both appear to work equally well, in practice the proximal user might be inclined to choose one over the other based upon which manipulated video they find more appealing or realistic. 

Although our results did not appear to be skewed, there was a discrepancy in the replace video manipulation that was used during the Hallway scene. In all other video clips the filters were placed over the entire door frame to fully block out the hallway. However, in using the replace video manipulation a filter was placed only over the cardboard box. Our main concern with this was that there might be more privacy violations since participants could still mention other details about the hallway (e.g. the stairs, banister, exit sign). Since no privacy violations occurred, it can be deduced that replacing only the box was enough to take attention away from the hallway, even though the question asked about it directly. 

 It is also hypothesized that if the filter had been placed over the entire doorway the view through the doorway would not have been as clear and understandable. It is likely that the replace video manipulation would have given the appearance that there was a second door, and the user may have been confused by what they saw in the image. For this reason, a redact video manipulation would be preferable because it gives an obvious cue to the user that there is something beyond the doorway, but they are not allowed to view it. 

\subsubsection{Can't Discern}
 Blur was not significantly different than redact or replace, but it did allow some privacy violations. One possible reason for this could be due to the intensity of the blur filter. It is possible that a blur filter with a higher intensity would better protect the titles of the novels. Another factor that played a role in identifying the filtered books was the fame and notoriety associated with these particular books. With their iconic cover art, it is possible the participant could deduce which book was being filtered out as long as they could still get a sense of what the illustration on the cover looked like even if they could not read the title.

As far as privacy violations are concerned it is clear from our results that the blur, redact, and replace video manipulations work best at making privacy violations impossible. However, the ``Can't discern'' privacy type also expects that the user can still correctly identify the quantity of books on the shelf, including the filtered books. Not surprisingly, it seems that the more privacy violations that the video manipulation had on average, the better the video manipulation also did with correctly identifying the number of books on the shelf. It is important to note that in correctly recording the total number of books blur did do better than replace and redact, although not significantly. This indicates that blur may be better at balancing awareness and privacy protection. The replace filter had the lowest average number of books recorded on the shelf, which leads us to conclude that it may not be effective enough in giving the remote user the necessary awareness of the scene. At the same time, the redact filter is lacking in privacy protection because it does not have as subtle of an appearance as replace or blur. One can hypothesize that using a redact filter might draw unwanted attention to the region that is to be indiscernible, while a blur filter could be written off as simply bad video quality and would not be as alarming or confusing as a black box that block's the user's view. For this reason, we recommend using a blur filter, over redact or replace, for ``Can't discern'' privacy types.

The case for redact in this video clip is also very unique in that the boxes used to redact the books from the bookshelf were a very similar color to the bookshelf itself. This led to an effect that looked quite similar, although not identical, to replace. Perhaps if the boxes used had been a color such as red that created a greater contrast against the back drop of the bookshelf, or if the bookshelf itself had been a bright color to contrast with the boxes used for redacting, the results would more likely show a greater difference between results for the replace and redact video manipulations. Also, the shape of the boxes was similar enough to the shape of a standard book that it could be deduced that the boxes were hiding books as well. Varying the shapes of the redaction filter used could lead to less participants viewing the box as a representation for the book.  

\section{Task Performance Study}
\label{sec:taskPerfStudy}
Our study presented in Section~\ref{sec:vidManipStudy} showed that certain video manipulation techniques can be used to uphold privacy expectations.  However, the overall effectiveness of such a technique could be negated if its use prevents or significantly incumbers the ability of the remote user to accomplish a given task.  The aim of the study presented in this secton is to examine how the use of these privacy-protecting techniques impacts task performance.

\subsection{Method}
This study uses an independent-measures design with one control group and one experimental group.  Participants were asked to teleoperate a mobile robot through an unfamiliar home environment and to respond to a brief set of survey questions asking them to identify cleaning supplies and equipment contained in the home.  For the experimental group, the environment was physically modified to hide any evidence of children in the home.   After completing the task of identifying cleaning supplies, participants were asked whether or not they believed that children regularly visit this home.

%

\subsubsection{Participants}
30 participants were recruited through flyers distributed via email and posted on bulletin boards in the local community.  Participants were compensated ten dollars for their participation.  Participants were told that they would be expected to, ``drive a robot around an apartment and answer a brief set of survey questions.''  The average time spent per participant, including training and answering all survey questions, was between 30-45 minutes.    

Basic demographic information was collected for each participant.  12 males and 18 females participated in the study.  The mean age of the participants was 28.  33\% of the participants were students, and 30\% of the participants played video games more than once a month.  Only one of the participants reported any familiarity with the apartment building in which the home environment was staged, and zero expressed an expert level of familiarity with robots or other remotely operated devices.

\subsubsection{Procedure}
\label{redactionProcedure}
Participants were asked to teleoperate a TurtleBot 2 using a PS3 joystick through an unfamiliar home environment.  Participants sat at a table with two laptops.  One laptop showed a full screen live video feed from the robot.  The other laptop was used for displaying a web page that provided the instructions to the participant and allowed them to answer the survey questions.  The participants were provided with a manually drawn floor plan of the apartment that they would be moving the robot through, but they were never allowed to see this space with their own eyes.  They only observed the space through the video feed from the robot.

Before the experiment began, participants were given a brief training session on teleoperating the TurtleBot.  They were asked to use the joystick to drive a TurtleBot that was in the room with them.  After gaining some initial familiarity with the controls, they were asked to navigate around an obstacle without looking at the robot, but only looking at the video feed on the laptop screen.

Once the training was completed successfully, participants were presented with the following prompt:

\begin{quote}
You work for a Home Cleaning Agency, and a new customer is asking what it would cost for your company to have someone come in and clean part of his home. To give him your best estimate of what it would cost, you would like to take a look around his home, and he has allowed you to do this through the use of his home robot. You will be able to operate it remotely and see through a live video what the house looks like. Your task is to take an inventory of the cleaning supplies and equipment including their locations and if possible their brands.

Your company gives a discount to customers who have their own cleaning supplies and equipment that they allow the company to use. This cuts down on costs since the company does not need to bring extra supplies or use up their own resources. We have included instructions and questions to help you identify if there are any cleaning supplies and equipment that this customer owns and may allow the cleaning company to use.
\end{quote}

Participants were then instructed to navigate first to the kitchen and then to the living room and to identify any cleaning supplies and equipment that they found there, including their specific locations and brand names.  The cleaning supplies included Bon Ami cleaning powder, Dawn dish soap, Pine Sol floor cleaner, and Bounty paper towels in the kitchen, and a Dirt Devil vacuum cleaner and a dust pan and brush in the living room.  

After completion of the task, a post-survey was administered that included the following question:

\begin{quote}
Based on what you saw, to what degree do you agree with the following statement: 

\textit{Children regularly visit this apartment.}  

[Strongly disagree, Disagree, Niether agree nor disagree, Agree, Strongly agree]
\end{quote}

\subsubsection{Privacy Protection Independent Variable}
Each participant was randomly assigned to the control or experimental group.  All participants 
completed the same task as described in Section~\ref{redactionProcedure}.  The only difference was that for participants in the experimental group, the environment was physically modified to hide any evidence of children in the home.

Evidence that might convey that children were regularly in the home included an umbrella stroller in the kitchen, toys in the living room, family photos on the coffee table in the living room, a baby gate in the doorway to the bedroom, and a walker toy in the bedroom.  For the experimental group, appropriate techniques were applied to hide these pieces of evidence.  The specific techniques were chosen based on the results from our video manipulation study (see Section~\ref{sec:vidManipStudy}) and applied physically to the environment, rather than in software, in order to ensure that no shortcoming of the software implementation would impact our results.  The privacy expectations and the techniques used to satisfy those expectations are described below.

\begin{itemize}
\item A \textbf{``Can't tell''}  privacy type was applied to the toys in the living room and the stroller in the kitchen.  To mimic a perfect replace filter, these items were removed from the environment.  
\item A \textbf{``Can't observe''} privacy type was applied to the bedroom.  To mimic a perfect redact filter, a black bed sheet was hung over the bedroom doorway.
\item A \textbf{``Can't discern''} privacy type was applied to the family photos on the coffee table.  To mimic a perfect blur filter, the photos in the picture frames were replaced with copies that had been blurred using image editing software.
\end{itemize}

\begin{figure*}
\centering
\includegraphics[width=1.0\linewidth]{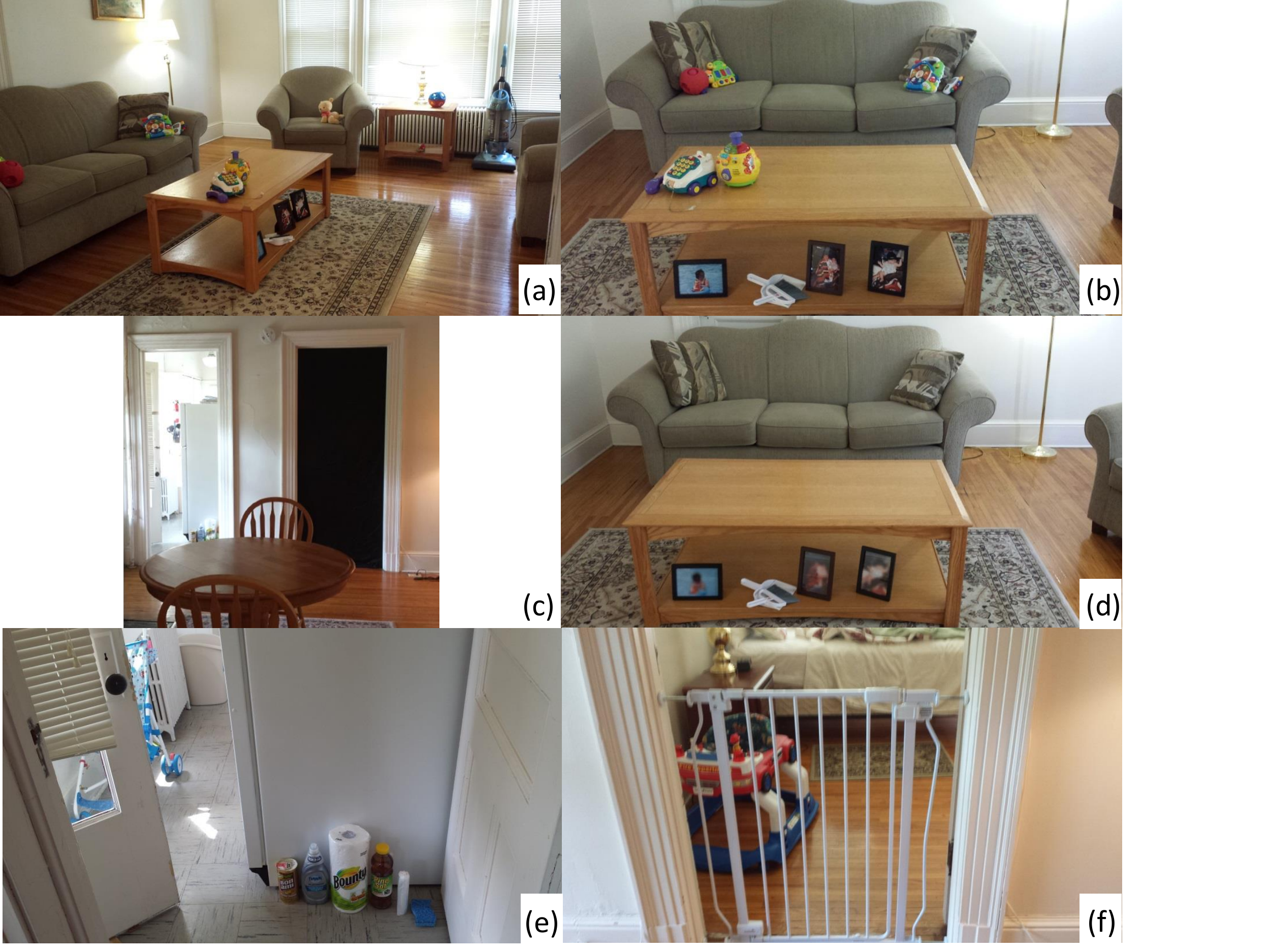}
\caption[Photographs of the home environment]{Photographs of the home environment.  Cleaning supplies in the living room included a vacuum cleaner (a) and a dust pan and brush under the coffee table (b).  For the experimental group, a sheet was used to black out the bedroom (c), the toys were removed, and the family photographs were blurred (d).  Additional cleaning supplies and a stroller were in the kitchen (e), and a baby gate and walker toy were in the bedroom (f).  }
\label{fig:collinsHouse}
\end{figure*}

Photographs of the home environment and videos from the user study are provided in ~\ref{fig:collinsHouse}. 

\subsubsection{Measures}
Our primary data is the participants' responses to the survey questions.  We also recorded all the sensor data from the robot, including the video stream, and we recorded audio of the participant talking aloud while completing the tasks.

\paragraph{Task Performance}
The participant's task was to identify cleaning supplies in the apartment.  Participants were presented with a list of 12 cleaning supplies and asked to check boxes next to the items that they found in the apartment.  For each participant, we recorded the number of correctly identified cleaning supplies.  We also recorded the number of incorrectly identified cleaning supplies, i.e., cleaning supplies checked on the list that were not actually present in the apartment.

For each cleaning supply identified, the participant was also asked to give its specific location in the apartment as well as its brand name.  We recorded the number of correctly identified locations and brands.

\paragraph{Privacy Violation}
As part of the post-survey questions, participants were asked to what degree they agreed with the statement that children regularly visit the apartment on a 5-point Likert scale.  As the privacy goal was to hide any evidence of children in the home, we use the participant response to this question as a measure of the degree to which privacy was or was not violated.

\subsection{Results}

\subsubsection{Task Performance}
\begin{figure}[h]
\centering
\includegraphics[width=1.0\linewidth]{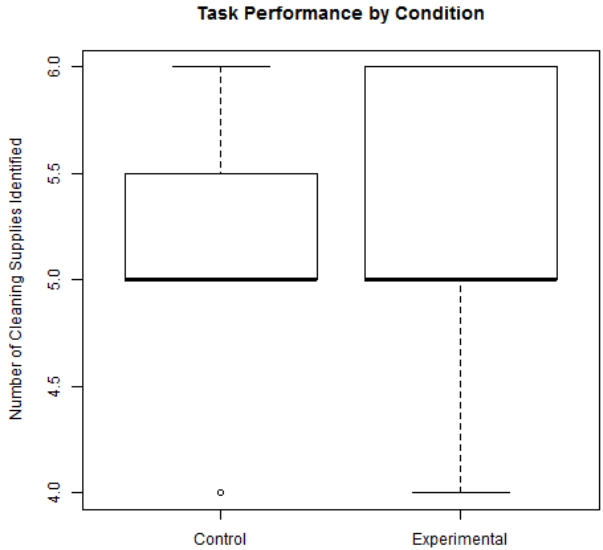}
\caption[Boxplot of the number of correctly identified cleaning supplies, by condition]{Boxplot of the number of correctly identified cleaning supplies, by condition.}
\label{fig:boxPerformance}
\end{figure}

A boxplot of the numbers of correctly identified cleaning supplies is shown in Figure~\ref{fig:boxPerformance}.  According to a Fisher's exact test ($p = 0.763$), there was no significant difference between the number of correctly identified cleaning supplies by participants in the control condition (mean of 5.13) and participants in the experimental condition (mean of 5.33).  Furthermore, there was also no significant difference found between the two conditions for the mean number of incorrectly identified supplies (0.13 control, 0.07 experimental), mean number of correctly identified locations (2.27 control, 2.40 experimental), or mean number of correctly identified brands (4.73 control, 4.40 experimental).  These results provide support for H2.

\subsubsection{Privacy Violation}
\begin{figure}[h]
\centering
\includegraphics[width=1.0\linewidth]{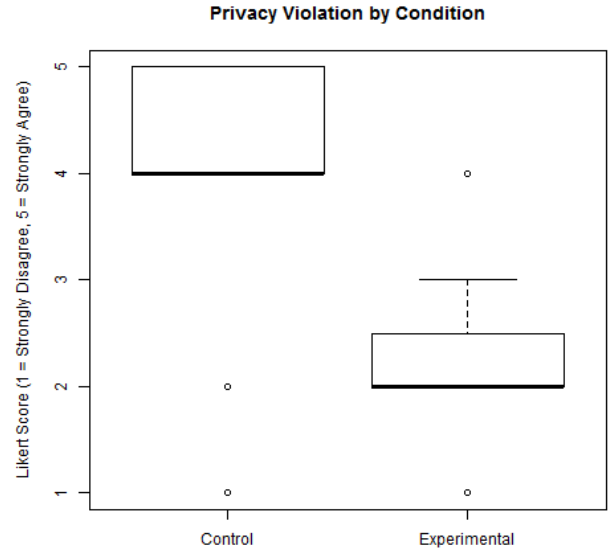}
\caption[Boxplot of the scores on a 5-point Likert scale in response to the statement that children regularly visited the apartment, by condition]{Boxplot of the scores on a 5-point Likert scale in response to the statement that children regularly visited the apartment, by condition.}
\label{fig:boxPrivacy}
\end{figure}

A boxplot of the scores on a 5-point Likert scale in response to the statement that children regularly visited the apartment is shown in Figure~\ref{fig:boxPrivacy}.  According to a Welch's t-test ($p < 0.001$), participants in the control condition (mean score of 3.86) agreed significantly more with the statement that children regularly visited the apartment than participants in the experimental condition (mean score of 2.13) where privacy-protecting techniques were applied.  This result provides support for H1.

\subsection{Analysis}
We studied the impact of privacy-protecting techniques on task performance in remote presence systems.  The results showed that the use of these privacy-protecting techniques did lead to fewer violations of privacy expectations without lowering task performance.

\subsubsection{Task Performance}
Participants were charged with the task of identifying cleaning supplies.  There were 6 supplies in the apartment, and participants in both conditions correctly identified a little above  5 supplies on average.  By far, the most common mistake among participants in both conditions was omitting the dust pan and brush.  This item was more difficult to identify due to its location under the coffee table.  Consequently, of the three places where supplies were located, this was also the most commonly omitted location.

Of the 6 supplies, all but the dust pan and brush had a brand name.  The brand name of the Dawn dish soap was the most difficult to read, and this was the brand that was most commonly missed by participants in both conditions.  Occasionally, this difficulty caused a participant to misidentify the supply as a bottle of bleach, rather than dish soap.  No participant in either condition listed a location or brand that was not actually present.

\subsubsection{Privacy Violation}
Using the privacy-protecting techniques did lead to fewer violations of privacy expectations, and this effect may be even more pronounced than we were able to measure.  Of the 15 participants in the control condition, only 3 disagreed with the statement that children regularly visited the apartment.  Interestingly, 2 of these participants seem to have clicked the wrong button because when asked to justify their answer, they gave reasons such as ``stroller in the kitchen, toys in the living room.''  Similarly, the one and only participant in the experimental group that agreed with the statement did so because, ``The apartment is really clean, I could not see any family pictures or toys around.''

\section{Discussion}
We have conducted two user studies to examine the protection of privacy in remote presence systems.  The results imply that video manipulation techniques are effective in obscuring details to protect privacy expectations. Based on our analysis of the collected data, we recommend using a replace video manipulation technique for ``Can't tell'' privacy types, a redact video manipulation technique for ``Can't observe'' privacy types, and a blur video manipulation technique for ``Can't discern'' privacy types. Furthermore, multiple techniques can be used simultaneously to uphold different privacy expectations without impacting the remote user's ability to perform a task.

\paragraph{Limitations}
In the video manipulation study each scene was manipulated using all five video manipulations, but there was only one scene used for each privacy type. To help generalize these findings, one could experiment with different conditions (e.g. lighting condition, number of objects/focal points in a scene, settings or calibrations for the video manipulations) to create more video clips and better analyze which video manipulation would be most appropriate for most scenes that attempt to uphold a certain privacy type.

 There are also other kinds of video manipulations that were not used in this study that could be used and tested to see if they would do any better with upholding certain privacy types. Even the video manipulation techniques used in this study could be modified further to better uphold privacy, such as using a stronger blur filter or a more robust inpainting technique for the replace filter. This study also made mention of how some video manipulations might be preferable for certain privacy types due to the appearance they gave to the remote user, and further investigation could be done to examine which type the proximal user prefers to show the remote user when multiple video manipulations are equally effective in protecting privacy. 

In the task performance study we observed our results in one environment, with one robot, one task, one set of privacy expectations, and one set of techniques to uphold them.  For this reason, we can only speculate about the generalizability of our observation that using privacy-protecting techniques reduces privacy violations without reducing task performance.  There are certainly scenarios where this may not be the case, or where different techniques may be needed.

Another limitation of this study is that we implemented the techniques by physically altering the environment, rather than manipulating the live video stream in software.  Indeed, this was an intentional part of our design, in order to avoid any effects that might be due to limitations of the software system.  We also minimized the effects of network lag by putting the robot and the machine used to teleoperate it on the same local wireless network.  Our goal in this study was to remove such potentially confounding variables, but these are important concerns that will need to be addressed in a real-world system.

\section{Conclusion}
\label{sec:conclusion}
Manipulations of the video stream can be used to protect some of the observational privacy concerns raised by the use of remote presence systems.  Certain video manipulation techniques are more effective than others for specifc privacy types:  replace for ``Can't tell'', redact for ``Can't observe'', and blur for ``Can't discern''.  These manipulations can be used in concert with one another to protect a given set of privacy expectations without interfering with the remote user's ability to accomplish a task.

\bibliographystyle{abbrv}
\bibliography{privacy}  

\appendix
\section{Coding Scheme}
To make sense of the recorded data we received it was necessary that the data be sorted and categorized, particularly into which answers should be recorded as privacy violations and which should not. As the data was analyzed guidelines were put into place on which responses would be considered privacy violations and which would not. In the rare cases where answers were completely unintelligible or could not be classified, the response was ommitted. 

For the ``Can't tell'' privacy type, privacy violations were recorded for any correct identification of one of the four valuables. Between the two coders, there was a .84 inter-rater reliability as based on Cohen's $\kappa$. One privacy violation was recorded for any mention of a television or LED screen. Another one privacy violation was recorded for any mention of a desktop computer, computer, or CPU. Up to two privacy violations were recorded for any mention of laptops or personal computers. The number of privacy violations for laptops was dependent on the number of laptops mentioned in the recorded text, and then further relied upon the total number of valuables recorded when the text was unclear. The ideal number of privacy violations is zero for this privacy type. It is important to note that any mention of a monitor was not counted as a privacy violation for the television, but was recorded as a privacy violation for the desktop if no other mention of the desktop was made. Also, many of the responses had more items listed than the valuables that were considered privacy violations. A frequent object listed was the desktop/monitor/computer/laptop on the far left that was actually just a painting on a shelf leaning against a wall. 

In the ``Can't observe'' privacy condition scenario, data was recorded about what, if anything, participants could see down a hallway. Among the different coders, there was a .94 inter-rater reliability as based on Cohen's $\kappa$. Most typically a privacy violation was recorded when any mention of the cardboard box was made. This box was also referred to several times as something being made of wood, a carton, and a package. An atypical response that was considered a privacy violation was the rare instance where the stairs were heavily implied in the recorded response when the participant mentioned that this was recorded on the second floor of a house. Since this answer demonstrated that the participant must have taken note of the stairs, which are in the hallway, we considered this a privacy violation. 

There were a good deal of commonly recorded items that were not seen as privacy violations since they were not located in the hallway, but rather inside the room: a chair, folding tables, a door stopper, the door itself, and the carpet.  It is important to note that any mention of a door was interpreted as a reference to the main wooden door that opens to the hallway. It was unclear whether any of the responses were referring to the door shape that could be seen at the end of the hallway, and due to the lack of specificity in responses it was determined that there was a greater likelihood that participants would have been referring to the wooden door.

 Another pertinent detail in recording privacy violations was that we did not consider any mention of a black box as a privacy violation when the redact video manipulation was being used because the assumption was made that this was referring to the filter instead of the cardboard box. This black box was also referred to as a black mass, black screen, and stone. While rare, there were a good number of items that were identified in participants answers that could not be traced back to anything in the video clip. Examples include an inverted cup, a glass bowl, a sofa, and a metallic button, and none of these were considered privacy violations since they did not exist or resemble anything that did exist in the hallway. 

 With video clips concerning the ``Can't discern'' privacy type, a privacy violation was recorded once for any mention of ``Twilight'' and once for any mention of ``Fifty Shades of Grey.'' There was a .98 inter-rater reliability as based on Cohen's $\kappa$. It should be noted that spelling did not have to be exact, but rather as long as the title was fairly close or had any of the main words from the title (i.e. ``fifty'' ``shades'' ``grey'' ``twilight'') the privacy violation was recorded. However, there was a unique case where a participant incorrectly identified a title as ``Breaking Dawn,'' which is another novel in the ``Twilight'' saga, and this was not recorded as a privacy violation because the title was clearly a guess and incorrect at that. In assessing accuracy in reading the unfiltered titles a correctly identified book title was recorded for any mention or relatively close spelling to ``ARToday'' and another was recorded for any mention of any of the words in the title or relatively close spellings to ``Visualization: The Second Computer Revolution.'' For title accuracy, there was a .98 inter-rater reliability as based on Cohen's $\kappa$.
\end{document}